\documentclass[11pt,letterpaper]{article}
\usepackage{emnlp2017}
\usepackage{times}
\usepackage{latexsym}
\usepackage{CJK}
\usepackage{url}
\usepackage{array}
\usepackage{graphicx}
\usepackage{multicol}
\usepackage{multirow}
\usepackage{float}
\usepackage{epsfig,epstopdf}

\emnlpfinalcopy

\def\emnlppaperid{***}

\title{Memory-augmented Neural Machine Translation}

 \author{Yang Feng$^{2,1,4}$, Shiyue Zhang$^{1,3}$, Andi Zhang$^{1,3}$, Dong Wang$^{1*}$ and Andrew Abel${^5}$  \\
 $^1$Center for Speech and Language Technologies, RIIT, Tsinghua University \\
 $^2$Key Laboratory of Intelligent Information Processing, \\ Institute of Computing Technology, Chinese Academy of Sciences \\
 $^3$Beijing University of Posts and Telecommunications, China \\
 $^4$Huilan Corporation, Beijing, China\\
 $^5$Xi'An Jiaotong Liverpool-University, Suzhou, China \\
 fengyang@ict.ac.cn, \{byryuer, andizhang912\}@gmail.com \\
 wangdong99@mails.tsinghua.edu.cn, andrew.abel@xjtlu.edu.cn
 }

\date{}

\begin{document}

\maketitle

\begin{abstract}

Neural machine translation (NMT) has achieved notable success in recent times, however it is also widely recognized that this approach has limitations with handling infrequent words and word pairs. This paper presents a novel memory-augmented NMT (M-NMT) architecture, which stores knowledge about how words (usually infrequently encountered ones) should be translated in a memory and then utilizes them to assist the neural model. We use this memory mechanism to combine the knowledge learned from a conventional statistical machine translation system and the rules learned by an NMT system, and also propose a solution for out-of-vocabulary (OOV) words based on this framework. Our experiments on two Chinese-English translation tasks demonstrated that the M-NMT architecture outperformed the NMT baseline by $9.0$ and $2.7$ BLEU points on the two tasks, respectively. Additionally, we found this architecture resulted in a much more effective OOV treatment compared to competitive methods.


\end{abstract}

\begin{CJK}{UTF8}{gbsn}
\section{Introduction}
\label{sec:intro}

Neural Machine Translation (NMT) has been shown to have highly promising performance, particularly when a large amount of training data is available~\cite{wu2016google,johnson2016google,mi2016coverage}.  Although there are different model architectures~\cite{sutskever2014sequence,Bahdanau:2014}, the common principle behind the NMT approach is the same: encoding the meaning of the input into a concept space and performing translation based on this encoding.  This `meaning encoding' principle leads to a deeper understanding and learning of the translation rules, and hence a better translation than conventional statistic machine translation (SMT) that considers only surface forms, i.e., words and phrases~\cite{koehn2003statistical}.

Despite positive results obtained so far, a particular problem of the NMT approach is that it has a tendency towards overfitting to frequent observations (words, word co-occurrences, translation pairs, etc.), but overlooking special cases that are not frequently observed. For example, NMT is good at learning translation pairs that are frequently observed, and can make use of them well at run-time, but for low-frequency pairs in the training data, the system may `forget' to use them when they should be.  Unfortunately, rare words are inevitable for all translation tasks due to Zipf's law, and indeed they are often the most important parts of a sentence, e.g., domain-specific entity names.  Table~\ref{tab:drift} shows an example, where the word `染色体( chromosomes)' is an infrequent word. As the system does not know (or has effectively `forgotten') this keyword, it does not translate
correctly, and an irrelevant translation is produced, leading to the phenomenon of `meaning drift'.  This weakness with regard to infrequent words/pairs with NMT has been noticed by a number of researchers, and some studies have been conducted to address this problem, e.g.,~\citet{luong2014addressing,cho2014properties,
li2016towards,Arthur:16,bentivogli2016neural,Zhangjiyuan:17}.

\begin{table}
\centering
\resizebox{\columnwidth}{!}{
\begin{tabular}{ll}
\hline
{\em src.} & 人类共有二十三对染色体。 \\
{\em ref.} & Humans have 23 pairs of chromosomes. \\
{\em NMT} &  There are 23-year history of human history.\\
\hline
\end{tabular}
}
\caption{An example of Chinese-to-English `meaning drift' with NMT. }
\label{tab:drift}
\end{table}

Superficially, this problem appears to be caused by the imperfect embeddings of infrequent words or the limited vocabulary size of NMT systems, but we argue that the deeper reason should be attributed to the nature of neural models: the translation function, represented by various neural networks, is shared amongst all of the translation pairs, so high-frequency and low-frequency pairs impact each other by adapting their shared parameters. Due to the overwhelming proportion of high-frequency pairs in the training data, the resulting trained model will naturally be much more focused on these frequently observed pairs.  More seriously, because the translation function is smooth, infrequent pairs tend to be wrongly seen as noise in the training process and so are largely ignored by the model.

In contrast to this, the conventional SMT approach is based on statistics of words and/or phrases, which, in principle, is a symbolic method that uses a discrete model and involves little parameter sharing.  The discrete model means that no matter how infrequently a pair occurs, its probability cannot be smoothed out, and the lack of shared parameters means that the frequent words or pairs have much less impact on infrequent words or pairs. Essentially, SMT memorizes as many of the observed patterns as possible, usually using a phrase table.

The respective advantages of SMT and NMT suggest that neither the pure neural approach nor the pure symbolic approach can provide a complete solution for machine translation, and a combined system that exploits the advantages of both approaches would be ideal.  This idea has been adopted in early research into neural-based MT methods, where neural models were utilized to improve SMT performance~\cite{zhang2015deep}. However, this seems to be counterintuitive, as intuitively learning general rules should be the first step, rather than first memorizing special cases and then learning general rules.  This suggests that the combined system should be primarily based on the neural architecture, with symbolic knowledge as a complementary support.

This paper presents such a neural-symbolic architecture, which involves a neural model component to deal with frequently seen patterns, and a memory component to provide knowledge for infrequently
used words and pairs.  More specifically, each memory element stores a source-target pair, specifying that a word defined by the source part should be translated to the word defined by the target part. This knowledge is then used to improve the neural model.  This is analogy to an experienced translator, who can work well in most cases using their own knowledge (i.e. the neural model aspect), but for unfamiliar and uncommon words that they have little experience of, they will still need to refer to a dictionary (i.e., the memory). This proposed memory-augmented NMT, or M-NMT, is therefore arguably much more similar to human translators than either NMT or SMT. 

\section{Attention-based NMT }
\label{sec:atten}

		\begin{figure*}[!h]
			\centering
			\includegraphics[width=1.5\columnwidth]{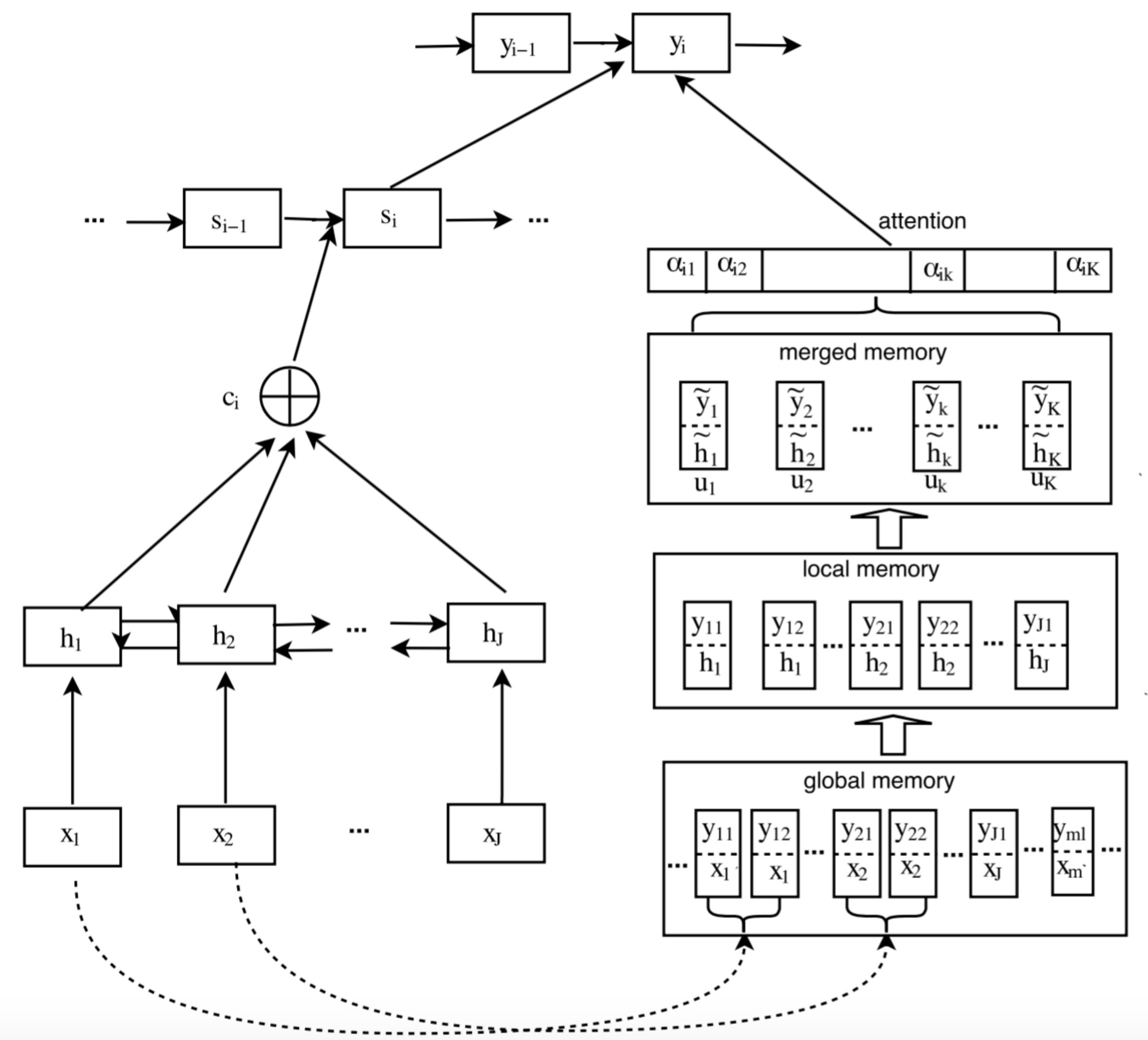}
			\caption{The structure of the M-NMT architecture.}
			\label{fig:model}
		\end{figure*}	
		
Before introducing our M-NMT architecture, we will give a brief review of our implementation of the attention-based RNN model first presented by~\citet{Bahdanau:2014}. This model is regarded as the state-of-the-art model and will be used as the baseline system in this study. Additionally, the neural model component of the M-NMT architecture uses the same attention-based RNN model, as being presented in the following.
		
The attention-based RNN model is based on an encoder-decoder frame, where the input word sequence $[x_1,x_2,...]$ in the source language is embedded as a sequence of hidden states $[h_1,h_2,...]$ by a bi-directional RNN with GRU as the hidden units, and another RNN is used to produce the target sentence $[y_1,y_2,...]$.  To force the generation to focus on a particular segment of the input at each generation step, \citet{Bahdanau:2014} proposed an attention mechanism.  Specifically, when generating the $i$-th target word, the attention factor of the $j$-th source word (and its neighbors, precisely) is measured by the relevance between the current hidden state of the decoder, denoted by $s_{t-1}$, and the hidden state of the encoder at the $j$-th word $h_j$.  This can be calculated by any similarity function, but a multiple layer perceptron (MLP) is often used, given by:

		\[
		\alpha_{ij} = \frac{e_{ij}}{\sum{e_{ik}}};  \ \ \ \ e_{ij} = a(s_{i-1}, h_j)
		\]

\noindent where $a(\cdot,\cdot)$ is the MLP-based relevance function, and $\alpha_{ij}$ is the attention factor of $x_j$ at decoding step $i$. The semantic content that the decoder focuses on, i.e. attended content, is then derived by:

		\[
		c_i = \sum \alpha_{ij} h_j.
		\]
		
The decoder updates the hidden state with a recurrent function $f_d$, formulated by:

		\begin{equation}
		\label{eq:md-dec-s}
		s_i = f_d(y_{i-1}, s_{i-1}, c_i),
		\end{equation}
		
\noindent and the next word $y_i$ is generated according to the following posterior:
		
		\begin{equation}
		\label{eq:pmd}
		p(y_i) = \sigma(y_i^T W z_i)
		\end{equation}

\noindent where $\sigma(\cdot)$ is the softmax function, $W$ is a parameter matrix for word vector projection.
The intermediate variable $z_i$ is computed by a neural net with a single maxout hidden layer $g$, given by:

		\[
		z_i = g(y_{i-1}, s_{i-1}, c_i).
		\]

\noindent We used Tensorflow to implement this model, and the training recipe largely followed the seminal paper
of~\citet{Bahdanau:2014}.	
		
\section{Memory-Augmented NMT}
		
This section presents the M-NMT architecture. We first introduce the model, and then describe how the memory is constructed.
		
\subsection{The Architecture}
		
The M-NMT architecture is illustrated in Figure~\ref{fig:model}. It involves two components: the model and the memory components.  The model component is a typical attention-based RNN model as presented in Section~\ref{sec:atten}, which is regarded as being good at dealing with frequent words and pairs, and the memory component provides knowledge for infrequent words and pairs that are not easy for the neural model component to learn.  The outputs of the two components are combined to produce a final consolidated translation.
		
\subsection{Memory Elements}
		
We define each item of memory as a mapping from a word in the source language to its translation in the target language. If there are multiple translations for a word, then several of the best will be added to the memory according to the probability of the translation, until the maximum number of target words is reached. A memory element can be formally written by:

		\[
		u_{jl}=\left[\begin{array}{c} y_{jl} \\  x_{j}  \end{array}\right]
		\]

\noindent where $y_{jl}$ is the $l$-th translation of word $x_j$. This mapping will be saved as a memory element and will be used during translation. We refer to this memory as the \emph{global memory}, which is static during all the running time. The global memory is shown on the bottom-right of Figure~\ref{fig:model}.

To translate an input sentence, the memory elements the source words of which
are in the input sentence are selected to form a \emph{local memory}. 
This is shown in the right-middle of Figure~\ref{fig:model}. 
In order to include the context information in the local memory, the source part $x_j$ is replaced by
its annotation $h_j$:

		\[
		u_{jl}=\left[\begin{array}{c} y_{jl} \\  h_{j}  \end{array}\right]
		\]

\noindent A consequence of the source encoding is that if a source word occurs multiple times in the sentence, all the occurrences should be put into the local memory, with different $h_j$ to distinguish the context of each.  Finally, the local memory is compressed as follows. For each distinct target word $\tilde{y}_k$ in the local memory, all the elements with $\tilde{y}_k$ as the target are merged into a single element $u_k$, for which the source part is the average of the source part of all the elements to be merged, given by:
		
		\begin{equation}
		\label{eq:merge}
		u_{k}=  \left[\begin{array}{c} \tilde{y}_{k} \\  \tilde{h}_{k}  \end{array}\right] =  \left[\begin{array}{c} \tilde{y}_{k} \\  \sum_j p(x_j|\tilde{y}_k)h_{j}  \end{array}\right]; \ \ \ \forall  \tilde{y}_{k} \in \{y_{jl}\}
		\end{equation}

\noindent where $p(x_j|\tilde{y}_k)$ means the probability that $x_j$ is translated into $\tilde{y}_k$ and can be obtained from either a human-defined dictionary or the dictionary of an SMT system.

		
		

\subsection{Memory Attention}
		
In order to use the information stored in the memory to improve NMT, we need to pick up appropriate elements
from the local memory at each translation step.  A similar attention mechanism as in the neural model is designed.
Denote the attention factor of each memory element $u_{k}$ at each translation step $i$ by $\alpha^m_{ik}$,
and assume it is derived from a relevance function $e^m_{ik}$:

		\[
		\alpha^m_{ik} = \frac{e^m_{ik}}{\sum_{k=1}^K {e^m_{ik}}},
		\]

\noindent where $K$ is the number of target words in the merged memory. The relevance function can be changed,
but in this study, we use a simple design:
		
		\begin{equation}
            \label{eq:em}
    		e^m_{ik} = (v^m)^\top tanh(W_s^m s_{i-1}  + W_u^m u_{k} + W_y^m y_{i-1} )
		\end{equation}

\noindent where $tanh(\cdot)$ is the hyperbolic function, $s_{i-1}$ is the current state of the decoder of the neural model, and $y_{i-1}$ is the generated word in the previous step. The parameters of the memory attention mechanism include $\theta^m=\{v^m, W_s^m, W_u^m, W_y^m\}$, as defined in Eq.~\ref{eq:em}. 
			
The attention factor $\alpha^m_{ik}$ can be used in different ways,	here they are simply treated as the posterior to predict the next word to generate. Since the normalization is over all the target words in the local memory rather than the full vocabulary, treating $\alpha^m_{ik}$ as the posterior of all words is only an approximation, but was found in our experiments to be a good solution.  This memory-based posterior is combined with the posterior of the neural model, resulting in a consolidated posterior, given by:

		\[
		\tilde{p}(y_i) = \beta  \alpha^m_{ik} + (1-\beta) p(y_i)
		\]

\noindent where $p(y_i)$ is the posterior produced by the neural model, as shown in Eq.~\ref{eq:pmd}, and $\beta$
is a pre-defined interpolation factor.  Here $\alpha^m_{ik}$ corresponds to the attention to the same word in the merged memory as the predicted word $y_i$.
This simple posterior combination indicates the flexibility of the M-NMT architecture.  Existing knowledge can be compiled into the local memory to improve model-based prediction, or if no knowledge is available, the system  will rely on conventional NMT.
		
An advantage of this simple combination is that the memory component can be trained independently of the neural model.  		
We set the objective of the training is to let the memory attention as accurate as possible. Given the $n$-th training sequence,
at each translation step $i$, the target attention should be $1$ on the current word $y^n_i$ and $0$ elsewhere. The
objective function therefore can be written as the cross entropy between the target attention and the
output of the attention function, given as follows:

		\[
		  L(\theta) = \sum_n \sum_i log (\alpha^m_{i k_i^n})
		\]

\noindent where $k_i^n$ is the position of $y^n_i$ in the merged memory.
The optimization is conducted with respect to the parameters $\theta^m$.  The optimization algorithm is
the stochastic gradient descent (SGD) with AdaDelta to adjust the learning rate~\cite{zeiler2012adadelta}.
		
It should be noted that joint training of the memory and the model is possible, but it requires a large amount of
GPU memory and risks over-fitting. Therefore, we only train the memory, with the model component unchanged.  
Efficient model-memory joint training is beyond the scope of this paper and can be investigated in future work.
Particularly, the parameter $\beta$ could be optimized to balance the contribution from the model part and the 
memory part, but constrains have to be carefully settled to avoid overfitting to the training data.

\subsection{Memory for SMT Integration}
		
The M-NMT architecture is a flexible framework that provides extra knowledge to the conventional model-based
NMT. If the knowledge is generated by a conventional SMT system, it is essentially an elegant combination of SMT and NMT.  In this work, we use the translation dictionary produced by an SMT system as the knowledge to create the memory, which involves first aligning the training sentence pairs using the GIZA++ toolkit \cite{Och:03b} in both directions, and applying the ``intersection'' refinement rules \cite{koehn2003statistical} to get a single one-to-one alignment for each sentence pair, and then extracting the translation dictionary based on these alignments. We can see the dictionary as the phrase pairs of the length 1 and leave the phrase pairs longer than 1 as future work.

The key information provided by the dictionary is the conditional probability that a source and a target word are translated to each other.
This information is used twice during local memory construction. Firstly, the conditional $p(y_{jl}|x_j)$ is used to select the most possible target words $y_{jl}$ to participate the local memory, and secondly, the conditional $p(x_j|\tilde{y}_k)$ is used to merge the elements whose target words are $\tilde{y}_k$, as shown in Eq.~\ref{eq:merge}.

\subsection{Memory for OOV Treatment}
\label{sec:method:oov}
		
The memory also provides a flexible way to address OOV words.  OOV words can be defined in multiple ways, but
here we focus on true OOVs that are totally new in both bilingual and monolingual data (i.e. rare words that are not present in any training data). 
One example is when a model is migrated to a specific domain.  To address these OOVs, we firstly need a manually defined dictionary to specify how an OOV word should be translated, where the target word could be either an in-vocabulary word or an OOV.
		
This dictionary will be used as the knowledge to construct the local memory at run-time.  Specifically, if an OOV
word is encountered on either the source or target side during local memory construction, the vector of a similar word is borrowed to
represent the OOV word.  Since the words are totally new, the similar word has to be defined manually. To avoid any confusion with other words, the selected similar word should not appear in the existing input sequence if the OOV is in the source side,	and should not match any target words in the existing local memory. 
To achieve this, several candidates have to be pre-defined for each OOV, so that alternative choices are available at run-time.  A problem of this approach is that the vocabulary of the neural model is fixed, so cannot output probabilities for OOV words.
To solve this, we let the selected similar word entirely overwritten by the OOV word, and any prediction for the similar word will be `re-directed' to the OOV word.

\section{Related Work}

The idea of memory augmentation was inspired by recent advances in the neural Turing machine~\cite{graves2014neural,graves2016hybrid} and memory network~\cite{weston2014memory}.  These new models equip neural networks with an external memory that can be accessed and manipulated via some trainable operations.  The memory idea has been utilized in NMT. For example, \citet{Wang:16} used a memory to extend the state of the decoder RNN in the attention-based NMT. In this case, the contribution of the memory is to provide temporary variables to assist RNN decoding. In contrast, our work uses memory to store knowledge. The memory in \citeauthor{Wang:16}'s work could be considered to be note paper, while the memory in our work is more like a dictionary.
		
The idea of combining SMT and NMT was adopted by early NMT research, but these combinations were mostly based on the SMT framework, as discussed in depth in the review paper from~\citet{zhang2015deep}.  \citet{Cohn:16} proposed to enhance the attention-based	NMT by using some structural knowledge from a word-based alignment model. The focus of their work was to use the extra knowledge to produce a better attention. In contrast, our work promotes the target words directly using the word mapping stored in the memory.  \citet{Arthur:16} proposed to involve lexical knowledge to assist with translation, particularly for low-frequency words.  This is similar to our proposed idea, with the key difference that their work uses the attention information to select the target words, while ours trains a separate attention, based on both the source and target words.
		
Regarding handling OOV words, \citet{jean15onusing} presented an efficient training method to support a larger vocabulary, which
helps alleviate the OOV problem significantly. \citet{Stahlberg:16} used SMT to produce candidate results in the form of lattice and NMT to re-score the results. As SMT uses a larger vocabulary than NMT, some OOV words can be retained. 
\citet{sennrich2016neural} proposed a subword approach, where OOV words are expected to be spelled out by subword units.  \citet{luong2014addressing} proposed a post-processing approach that learns the position of the source word when an UNK symbol
is produced during decoding. By this position information, the UNK symbol (unknown words) 
can be replaced by the correct translation using a lexical table. \citet{li2016towards} proposed a replace-and-restore approach that replaces infrequent words with similar words before the training and decoding, and restores rare words and their target words, obtained from	a lexical table.  Compared to the work of \cite{luong2014addressing} and \cite{li2016towards}, which relies on post-processing, our M-NMT approach is more like pre-processing. This means that the required information for OOV words is prepared before decoding. This seems more flexible than the post-processing methods, as we can easily deal with multiple targets for OOVs, by letting the decoder select which target is the most appropriate.  Nevertheless, we do share the same idea of using similar words as in~\cite{li2016towards}, which we think	is inevitable if the OOV words are totally new.

\section{Experiments}

\subsection{Data}
		
The experiments were conducted for Chinese-English translation using two datasets, the relatively small IWSLT dataset, and the much larger NIST dataset. As we will see, the NMT and SMT approaches exhibit different behaviours on these two datasets, and the memory-augmentation approach offers different contributions to them. 
		
\vspace{2mm}
\noindent {\bf The IWSLT corpus}  The training data consists of 44K sentences from the tourism and travel domain. The development set was composed of the ASR devset $1$ and devset $2$ from IWSLT 2005, and testing used the IWSLT 2005 test set.
		
\vspace{2mm}
\noindent {\bf The NIST corpus}  The training data consists of 1M sentence pairs with 19M source tokens and 24M target tokens from the LDC corpora of LDC2002E18, LDC2003E07, LDC2003E14, and Hansard's portion of LDC2004T07, LDC2004T08 and LDC2005T06.  We use the NIST 2002 test set as the development set and the NIST 2003 test set as the test set.
		
\vspace{2mm}
\noindent {\bf Memory data} To construct the memory, we used the GIZA++ toolkit \cite{koehn2003statistical} to align the training data in both directions, and kept the word pairs that appeared in the phrase tables of both directions.  The global memory size is $80K$ for the IWSLT task, and $500K$ for the NIST task.  These word pairs were then filtered according to the conditional probability $p(w_t|w_s)$ where $w_s$ and $w_t$ are source and target language words, respectively.  For each $w_s$, at most two candidates of $w_t$ were retained.

\subsection{Systems}
		
We used a conventional SMT system and an attention-based RNN NMT system as the baselines, and investigated a variety of M-NMT architectures.
		
\vspace{2mm}
\noindent {\bf SMT baseline:} For the SMT system (denoted by {\em Moses}), Moses \cite{Koehn:07a}, a state-of-the-art open-source toolkit, was used. The default configuration was used where the phrase length was 7 and the following features were employed: relative translation frequencies and lexical translation probabilities on both directions, distortion distance, language model and word penalty. For the language model, the KenLM toolkit~\cite{Heafield:11} was used to build a 5-gram language model (with the Keneser-Ney smoothing) on the target side of the training data.
		
\vspace{2mm}
\noindent {\bf NMT baseline:} For NMT, we reproduced the attention-based RNN model proposed by~\citet{Bahdanau:2014}, which is denoted by {\em NMT}. The implementation was based on Tensorflow\footnote{https://www.tensorflow.org/}.  We compared our implementations with a public implementation using Theano\footnote{https://github.com/lisa-groundhog/GroundHog}, and achieved comparable
(even slightly better) performances on the same data sets with the same parameter settings.
		
\vspace{2mm}
\noindent {\bf M-NMT system:} The M-NMT system was implemented by combining the memory structure and the NMT system.  The model part is the same as the NMT baseline, while the attention function of the memory part was trained.  During the training, if the target word is an UNK symbol, or the target word is not in the memory (due to the limited word pairs in the memory), this word is simply skipped from back-propagation. This skipping is important as it avoids bias caused by the large amount of UNK symbols.  The trained M-NMT system can be readily used to deal with OOV words, without any re-training.
		
For M-NMT,  the complete form of the relevance function for memory attention is $e^m_{ik}(s_{i-1}, y_{i-1}, u_k)$. Of these, $s_{i-1}$ and $y_{i-1}$ are `attending factors' that represent the information used `to attend', while $u_k$, which consists of a source part $u_k(x)$ and a target part $u_k(y)$, involves `attended factors' that represent the content `to be attended'.  To investigate the contribution of different attending and attended factors, these factors are combined in	different ways, leading to different M-NMT variants, as shown in~\ref{tab:m-nmt-variant}. Note that {\em M-NMT($s; u^{y}$)} is the simplest configuration and the attention essentially learns a target-side language model. {\em M-NMT($s; u^{xy}$)} involves the source part of the memory, which implicitly learns a bilingual language model. Involving the decoding history $y_{i-1}$ makes this learning more explicit.
		
		\begin{table}[!t]
			\centering
			\scalebox{1}{
				\begin{tabular}{|l|l|l|}
					\hline
					System & Attending & Attended \\
					\hline
					M-NMT($s, u^{y}$)         &  $s_{i-1}$ & $u_k(y)$ \\
					M-NMT($s, u^{xy}$)        &  $s_{i-1}$ & $u_k(x), u_k(y)$ \\
					M-NMT($sy, u^{y}$)      &  $s_{i-1}, y_{i-1}$ & $u_k(y)$\\
					M-NMT($sy, u^{xy}$)      &  $s_{i-1}, y_{i-1}$ & $u_k(x), u_k(y)$\\
					\hline
				\end{tabular}
			}
			\caption{M-NMT systems with different configurations.}
			\label{tab:m-nmt-variant}
		\end{table}
		
\vspace{2mm}
\noindent {\bf Settings} For a fair comparison, the models configurations in the NMT system and the M-NMT system were intentionally set to be identical. The number of hidden units,  the word embedding dimensionality and the vocabulary size were empirically set to $500$, $310$ and $30000$, respectively.  In the training process, the batch size of the SGD algorithm was set to $80$, and the parameters for AdaDelta were set to be $\rho=0.95$ and $\epsilon=10^{-6}$. The decoding is implemented as a beam search, where the beam size was set to be $5$.
		
\vspace{2mm}
\noindent {\bf Evaluation metrics} The translation performance was evaluated using the BLEU score with case-insensitive n $\le$ 4-grams~\cite{Papineni:02}.
		
\subsection{SMT-NMT Integration Experiments}
		
		\begin{table}[!t]
			\centering
			\scalebox{1}{
				\begin{tabular}{|l|c|c|}
					\hline
					System               & IWSLT05 & NIST03 \\
					\hline
					Moses                & 52.5 & 30.6 \\
					\hline
					NMT                  & 43.9 & 31.3 \\
					\hline
					NMT-L                & 45.9 & 31.7    \\
					\citeauthor{Arthur:16}           &  &   \\
					\hline
					M-NMT($s, u^y$)      & 49.8 & 32.3  \\
					M-NMT($sy, u^y$)     & 50.7 & 32.5  \\
					M-NMT($s, u^{xy}$)   &   51.4   & 32.8  \\
					M-NMT($sy,u^{xy}$)   & {\bf 52.9} & {\bf 34.0}  \\
					\hline
				\end{tabular}
			}
			\caption{BLEU scores with different translation systems on the two Chinese-English translation datasets.}
			\label{tab:integration}
		\end{table}

In the first experiment, the M-NMT architecture combined SMT and NMT by using SMT to construct the memory to assist with NMT.  For comparison purposes, the lexical prediction approach proposed by~\cite{Arthur:16} was also implemented. This uses the phrase table produced by SMT to improve NMT. Our implementation is a linear combination, and for a fair comparison, the neural model part was kept unchanged. At each step $i$, the auxiliary probability provided by the lexical part is $P(y_i)= \sum_j \alpha_{ij} P(y_i|x_j)$, where $\alpha_{ij}$ is the attention weight from the neural model, and $P(y_i|x_j)$ is obtained from the phrase table.  This can be regarded as a simple memory approach, with memory attention borrowed from the neural model, rather than being learned separately.
		
Table~\ref{tab:integration} shows the BLEU results with different systems.  Firstly, it can be observed that with the small IWSLT05 dataset, the SMT outperforms the baseline NMT, but with the large NIST dataset, NMT outperforms SMT. This is unsurprising as neural models often need more training data.  Secondly, the results show that with both datasets, the lexical approach (NMT-L) can improve NMT performance, showing that using SMT knowledge helps NMT. However, the improvement seems less significant than reported in~\cite{Arthur:16}. This is likely to be because our implementation focuses on creating a simple, extensible, and generalizable system, and therefore does not allow re-training the neural model.
				
The M-NMT system provides significant performance improvement, even with the simplest setting (M-NMT($s,u^y$)). More information factors tend to offer better performance, and the best M-NMT system, M-NMT($sy,u^{xy}$), outperforms the baseline NMT by $9.0$ and $2.7$ BLEU points on the two datasets respectively.  Notably, the improvement with the IWSLT05 dataset is impressive, the best M-NMT system outperforms even the very strong SMT baseline, which strongly supports our conjecture that NMT must be equipped with a symbolic structure to deal with infrequent words. It also suggests that the M-NMT architecture is a promising way to apply neural methods to low-resource tasks.

\subsection{OOV Treatment Experiments}
		
		\begin{table}[!t]
			\centering
			\scalebox{0.9}{
				\begin{tabular}{|l|c|c|c|c|}
					\hline
					& \multicolumn{2}{c|}{T-INV} & \multicolumn{2}{c|}{T-OOV} \\
					\hline
					System       & Recall & BLEU  & Recall & BLEU\\
					\hline
					NMT                               & 0.06 & 15.1 & 0 & 13.7 \\
					\hline
					M-NMT                       & 0.05 & 16.0 & 0  & 14.6 \\
					\hline
					NMT-PL                            & 0.09 & 15.4 & 0.08 & 14.3  \\
					\citeauthor{luong2014addressing}                &&&& \\
					\hline
					M-NMT+OOV                             & {\bf 0.28} & {\bf 17.0}  & {\bf 0.40} &{\bf 15.9}  \\
					\hline
				\end{tabular}
			}
			\caption{The OOV recall rates and BLEU scores on sentences with OOV words. `T-INV' refers to the case where the target words of the OOV input are in-vocabulary, and `T-OOV' means the case where the target words are also OOV.}
			\label{tab:oov}
		\end{table}
		
Here the M-NMT architecture was used to handle OOV words.  The experiments were conducted on the NIST dataset, for which we collected $312$ test sentences containing OOV words. This test set was divided into two subsets: the T-INV set, containing sentences with source OOV words whose translations are NOT OOV in the target language; and the T-OOV set, containing sentences with OOV words that are OOV in both source and translation. There were $491$ source-side OOV words in total, among which $276$ words have in-vocabulary translations and $215$ words only have OOV translation. We constructed a translation table with three items for each OOV word: (1) its translation; (2) its similar word; (3) the similar word of its translation, if the translation is also an OOV word.  All the above was designed by hand, and for each OOV word, there was only a single translation.  Although it is not difficult to collect most of this information automatically (e.g., by using an SMT phrase table), we are simulating the scenario where OOV words are newly coined, or where the system is migrated to a new domain, meaning that some words are totally new to the system. Handling OOV words of this type is certainly challenging, but it is also practically valuable.
		
For comparison, the place-holder approach proposed by~\citet{luong2014addressing} was also implemented. Here, OOV words in the target language are substituted by position-aware UNKs, and a post-processing step is used to replace UNKs with the correct translation. We denote this system as `NMT-PL'. For M-NMT, only the best configuration M-NMT($sy,u^{xy}$) was tested in this experiment. Two scenarios were considered: the original M-NMT system, and the M-NMT system with OOV words involved in the memory (denoted by M-NMT+OOV).

		\begin{figure}[!t]
			\centering
			\includegraphics[width=0.9\columnwidth]{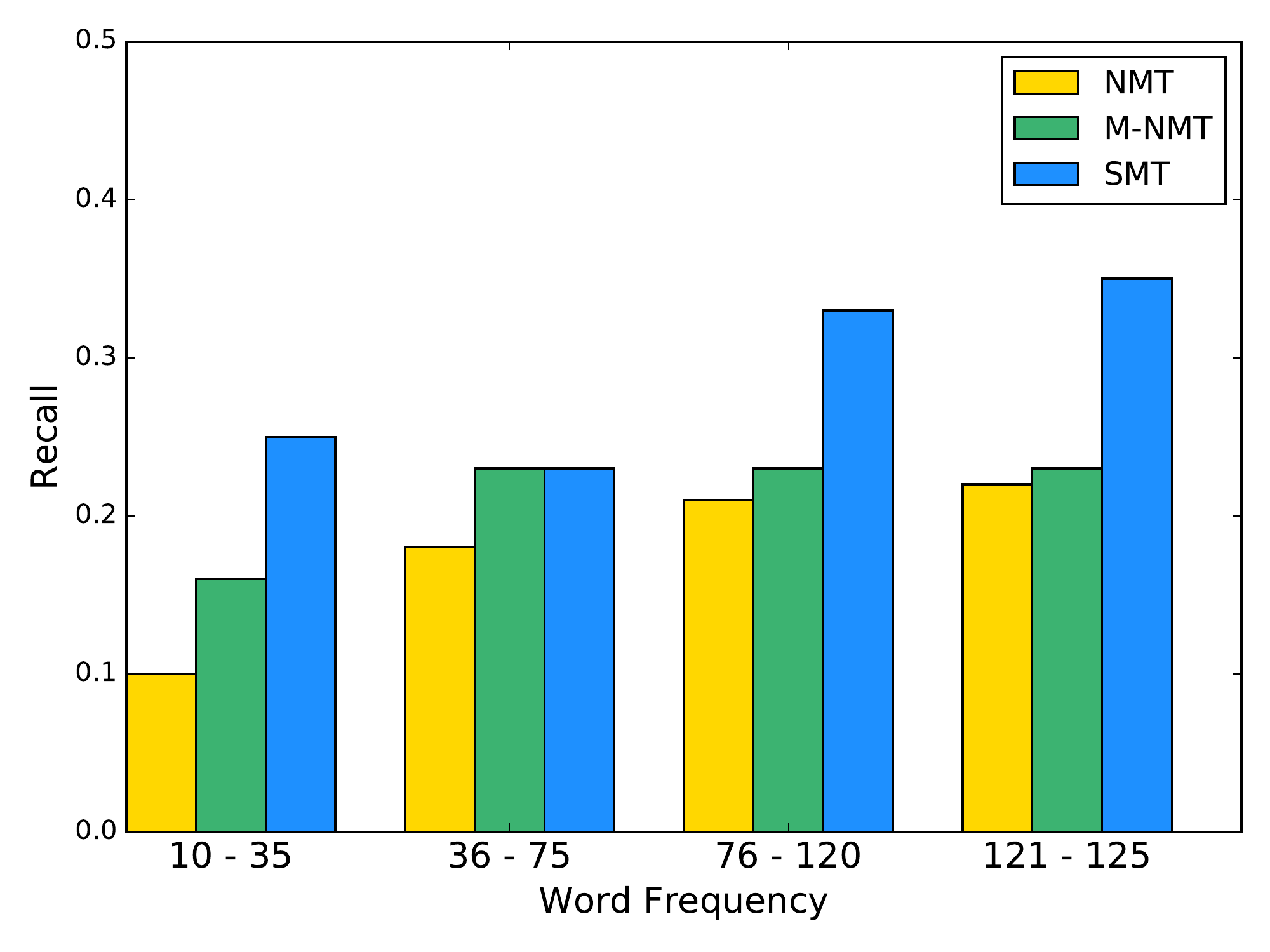}
			\caption{The recall rates of words in different frequency bins.}
			\label{fig:frequency}
		\end{figure}

		\begin{table}[!t]
			\centering
			\resizebox{\columnwidth}{!}{
				\begin{tabular}{ll}
					\hline
					{\em src.} & 人类共有二十三对染色体。 \\
					{\em ref.} & Humans have 23 pairs of chromosomes. \\
					{\em Moses} & A total of 23 human chromosome.  \\
					{\em NMT} & There are 23-year history of human history . \\
					{\em M-NMT} & There have a total of 23 species of chromosomes . \\
					\hline
				\end{tabular}
			}
			\caption{The translations from different systems for the Chinese-to-English `meaning drift' example.}
			\label{tab:eg}
		\end{table}

The results with the NIST dataset are shown in Table~\ref{tab:oov}. In addition to the BLEU scores, we also report the OOV recall rates, defined as the proportion of OOV words that are correctly translated. It can be seen that both the basic NMT and M-NMT systems work badly with OOV words: they can only process OOV words whose translations are not OOV in the target language, and the recall rate is very low ($0.05$ approximately). The place-holder approach (NMT-PL) can address both types of OOV words, but the recall rate is still low. The M-NMT system with OOV memory, in contrast, is much more effective in OOV word translation, as shown in Table~\ref{tab:oov}.  We also implemented the replace-and-restore approach reported by~\citet{li2016towards}, but found performance to be poor (the BLEU scores are $13.9$ on T-INV and $13.3$ on T-OOV). This may be due to no re-training of the neural model (again, in order to keep the extensibility and generalizability of the M-NMT system).

\subsection{Frequency Analysis}
		
To form a better understanding of the memory mechanism, we distribute the test sentences into four bins according to the lowest word frequency among the sentence and compute the recall rates for words in these bins. Once a generated translation word is also in one of the references, we treat it as one hit. The experiment was conducted with the NIST dataset.  The results in Figure~\ref{fig:frequency} show that M-NMT offers more improvement with infrequent words, in accordance with our argument that the memory mechanism helps NMT in dealing with infrequent words.

Finally, we demonstrate the translation with M-NMT for the example in Table~\ref{tab:drift}, as shown in Table~\ref{tab:eg}. It can be clearly seen that the memory has remembered the infrequent word `Chromosome', which resulted in an improved translation.

\end{CJK}

\section{Conclusions}
		
This paper presented a memory-augmented NMT approach, which introduces a memory mechanism for conventional NMT to assist with translating words not well learned by the neural model. Our experiments demonstrated that the new architecture is highly effective, providing performance improvement by $9.0$ and $2.7$ BLEU scores on two Chinese-English translation datasets, respectively. Additionally, it offers a very flexible and effective OOV treatment.  In our experiments, 
The OOV recalls are 28\% and 40\% for the OOV words whose target words are INV and OOV, respectively, a significant improvement on competing methods. Future work will investigate better model-memory integration, e.g., by joint training.

\section*{Acknowledgement}
This paper was supported by the National \mbox{Natural} Science Foundation of China (NSFC) under the project NO.61371136, NO.61633013, NO.61472428, NO.61472428.

\newpage
\bibliographystyle{emnlp_natbib}
\bibliography{allref,additional}

\end{document}